\documentclass[letterpaper, 10pt, conference]{ieeeconf}
\ifdefined\pdfsuppressptexinfo\pdfsuppressptexinfo=-1\fi
\ifdefined{/Title (GRAFT: Geometry-Aware Residual Adaptation for Foundation Trackers) /Author (Liang Li, Guoyao Zhang, Xingyu Chen, Sikai Wu, Qianyang Wu, Yuhang Lin, Renjing Xu, Mingliang Zhou, Diyun Xiang)}\fi
\IEEEoverridecommandlockouts
\usepackage{graphicx}
\usepackage{amsmath}
\usepackage{amssymb}
\usepackage{booktabs}
\usepackage{multirow}
\usepackage{array}
\usepackage{rotating}
\usepackage{xcolor}

\makeatletter
\newcommand{\nativefigurecaption}[1]{%
  \def\@captype{figure}%
  \caption{#1}%
}
\makeatother

\title{\LARGE \bf GRAFT: Geometry-Aware Residual Adaptation for Foundation Trackers}

\author{Liang Li$^{1}$, Guoyao Zhang$^{1}$, Xingyu Chen$^{1,2}$, Sikai Wu$^{1}$, Qianyang Wu$^{1}$, Yuhang Lin$^{1}$,\\
Renjing Xu$^{2}$, Mingliang Zhou$^{1\dagger}$, Diyun Xiang$^{1}$\\
{\normalsize \texttt{https://graft-humanoid.github.io/}}}

\begin{document}

\twocolumn[{%
\renewcommand\twocolumn[1][]{#1}%
\maketitle
\begin{center}
\centering
\includegraphics[width=\textwidth]{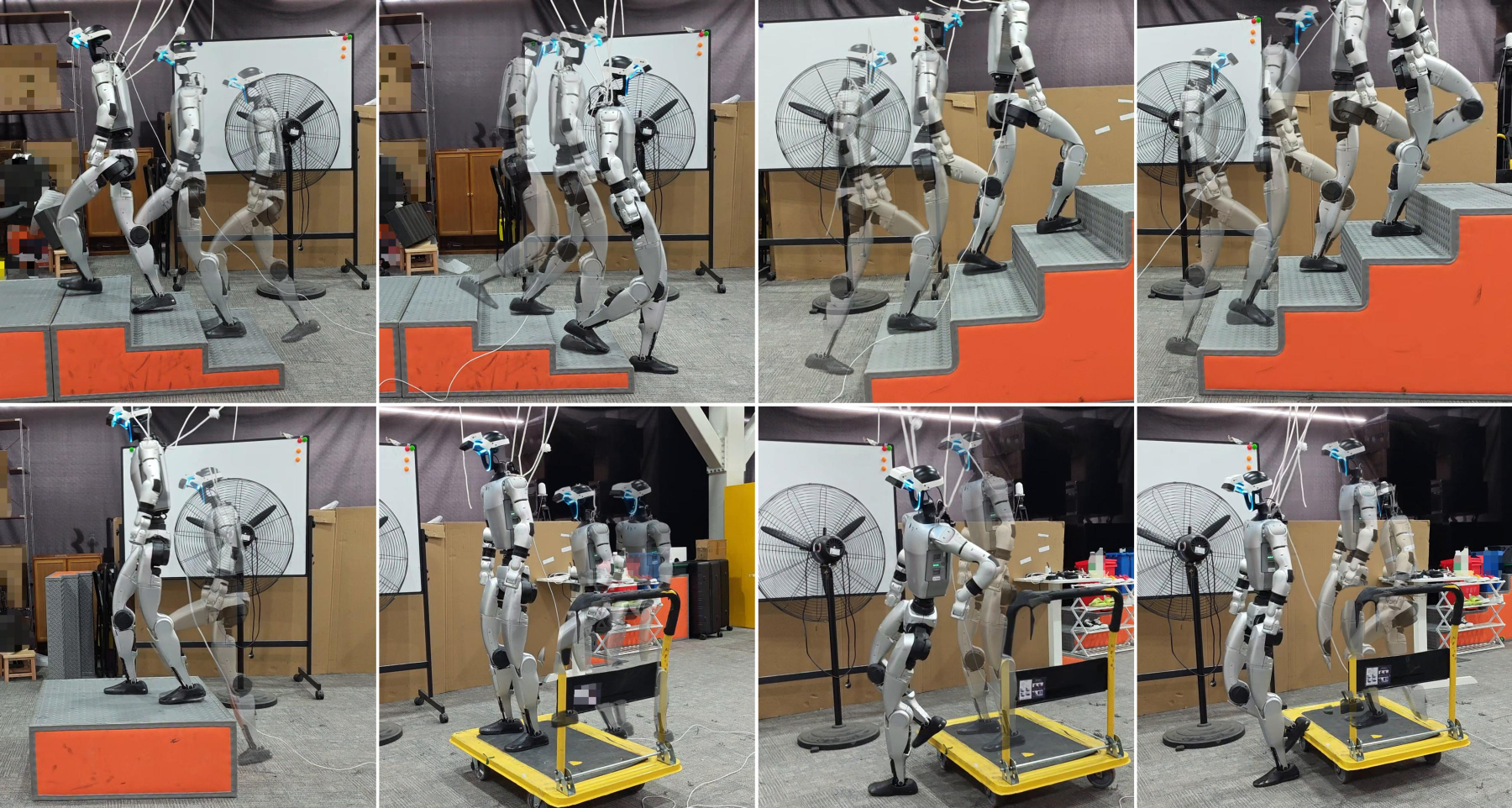}
\nativefigurecaption{GRAFT on hardware. The frozen SONIC base is driven by an unmodified reference while a
geometry-aware lower-body residual adapts execution to the terrain: a 0.30\,m platform, 0.10\,m and
0.17\,m steps, and a push-cart traversal. Frames within a scene are overlaid
with increasing opacity.}
\label{fig:teaser}
\end{center}
}]
{\renewcommand{\thefootnote}{}\footnotetext{$^{1}$Xiaomi Robotics Lab. $^{2}$HKUST(Guangzhou). $^{\dagger}$Corresponding author.}}

\begin{abstract}
Humanoid behavior foundation models (BFMs) provide reusable whole-body motion priors: given a kinematic reference, they generate coordinated joint commands for motion replay, teleoperation, and VLA control. However, trained primarily on flat-ground motion, BFMs lack an explicit mechanism to adapt foot placement, swing clearance, and contact timing to local terrain. Many terrain-aware BFM approaches address this limitation by retraining or fine-tuning the tracker together with perception, incurring substantial data and computation costs while coupling ordinary tracking to the perception pathway. To bridge this gap, we propose GRAFT, a geometry-aware residual adaptation framework for frozen behavior trackers. GRAFT preserves the BFM parameters and motion-reference interface, while a lightweight adapter adds terrain-conditioned corrections only to lower-body execution targets. The adapter is trained with task and foot-geometry rewards, regularized to suppress unnecessary corrections on flat ground, and distilled from privileged terrain observations to onboard depth. Because the frozen BFM already provides whole-body reference tracking and nominal locomotion, GRAFT reduces terrain learning to a focused execution-adaptation problem. We instantiate the framework on four frozen trackers and evaluate stairs, gaps, and platforms in simulation. We further validate it through real-world deployment on a Unitree G1, where the adapted SONIC tracker traverses terrain with onboard depth and unmodified references.
\end{abstract}

\section{Introduction}

A behavior foundation model (BFM) learns a single whole-body tracking policy from a large corpus of human motion: given a kinematic reference, it outputs joint commands~\cite{sonic,humanoidgpt,holomotion}. Motion references form its sole upstream command interface. The same BFM can therefore be driven by a recorded clip, an online teleoperator, or a high-level policy such as a vision-language-action (VLA) model, without retraining for each source. This shared reference interface is what lets one released tracker serve several upstream command sources.

These corpora, however, are recorded almost entirely on flat, empty ground, and the environment enters neither the policy's observation nor its reward. A BFM therefore learns how to approximate a reference kinematically, not how to place its feet on the surface it is actually crossing. Changing the reference does not supply what is missing. Given a clip of a person climbing stairs, the policy reproduces the motion in the air but does not actually climb the stairs, because the foot placement, swing clearance and contact timing the steps require were never part of what it was trained to produce.

Current approaches give BFMs scene-interaction ability by pairing motion with reconstructed scene geometry and retraining~\cite{gigabrain}, or by fine-tuning a pretrained model together with a terrain encoder~\cite{grail,pbfm}. These approaches are effective, but they change the deployed tracker and couple ordinary tracking to the terrain-perception pathway. Constructing terrain-paired motion data requires annotation and processing at scale, together with the compute to retrain on it. Fine-tuning a behavior prior on a narrow new task can also degrade the reusable prior, which motivates residual adaptation on a frozen backbone elsewhere~\cite{mosaic}.

To address this problem, we propose GRAFT. The frozen BFM processes the unmodified reference, while a separately trained adapter adds geometry-conditioned corrections to the lower-body execution targets; the upper-body targets remain those of the frozen tracker, since they are what a teleoperator or a VLA is commanding. A deployment gate controls whether the correction is applied at all.

The key idea is to separate reusable whole-body motion generation from terrain execution adaptation. The BFM retains responsibility for reference tracking, nominal locomotion, and whole-body coordination; the adapter learns only the terrain-dependent deviation from this nominal execution. We train it with foot-clearance and placement penalties and suppress its output on flat ground, then distill a teacher trained under privileged terrain observations into a student that reads onboard depth. This decomposition turns terrain adaptation into a focused execution problem without rewriting the upstream reference or modifying the released tracker.

Our contributions are:

\begin{itemize}
  \item We introduce GRAFT, a terrain-adaptation framework that adds geometry-conditioned lower-body corrections to a frozen behavior tracker while preserving its parameters and its motion-reference interface, with a gate that can be switched off at deployment; mid-run disabling and re-enabling are verified in simulation.
  \item We propose GRAFT as a general framework for adapting frozen behavior foundation trackers to terrain interaction. We instantiate the framework on four distinct trackers, training a separate adapter for each, and show that the same residual structure and reward family acquire terrain traversal while keeping flat-ground corrections small.
  \item We validate GRAFT through real-world deployment on a Unitree G1. The SONIC adapter is distilled from privileged terrain observations to onboard depth, and the resulting tracker traverses terrain with unmodified references; a separate composite-course simulation verifies mid-run residual disabling and re-enabling.
\end{itemize}

\section{Related Work}
\label{sec:rw}

\begin{table*}[t]
\centering
\caption{Capability comparison, grouped by the lines of work discussed above. Whole-body reference means the deployment command is a whole-body motion reference rather than a root velocity or a navigation goal; terrain sensing means that the deployed policy consumes robot-side exteroception; base intact means that no base parameter is modified; plug-and-play means that the residual can be disabled at deployment, returning the base action; multi-base means that one procedure is instantiated on two or more distinct base models. ``--'' marks a cell that does not apply, for instance the adaptation columns of a complete policy.}
\label{tab:capability}
\footnotesize
\setlength{\tabcolsep}{6pt}
\begin{tabular}{lccccc}
\toprule
Method & Whole-body reference & Terrain sensing & Base intact & Plug-and-play & Multi-base \\
\midrule
\multicolumn{6}{l}{\emph{Behavior foundation models}} \\
\quad SONIC~\cite{sonic}                     & \checkmark & $\times$   & --         & --         & --       \\
\quad Humanoid-GPT~\cite{humanoidgpt}        & \checkmark & $\times$   & --         & --         & --       \\
\addlinespace[2pt]
\multicolumn{6}{l}{\emph{Perceptive locomotion}} \\
\quad DPL~\cite{dpl}                         & $\times$   & \checkmark & --         & --         & --       \\
\addlinespace[2pt]
\multicolumn{6}{l}{\emph{Terrain-aware BFM}} \\
\quad GigaBrain-WBC~\cite{gigabrain}         & \checkmark & $\times$   & --         & --         & --       \\
\quad GRAIL, terrain branch~\cite{grail}     & \checkmark & \checkmark & $\times$   & $\times$   & $\times$ \\
\quad Perceptive BFM~\cite{pbfm}             & \checkmark & \checkmark & $\times$   & $\times$   & $\times$ \\
\addlinespace[2pt]
\multicolumn{6}{l}{\emph{Residual adaptation of a frozen policy}} \\
\quad ResMimic~\cite{resmimic}               & \checkmark & $\times$   & \checkmark & \checkmark & $\times$ \\
\quad MOSAIC~\cite{mosaic}                   & \checkmark & $\times$   & \checkmark & \checkmark & $\times$ \\
\quad Cerebellar residual~\cite{cerebellar}  & $\times$   & $\times$   & \checkmark & \checkmark & $\times$ \\
\midrule
\textbf{GRAFT (ours)}                        & \checkmark & \checkmark & \checkmark & \checkmark & \checkmark \\
\bottomrule
\end{tabular}
\end{table*}

\subsection{Whole-body motion tracking and behavior foundation models}
Whole-body motion tracking has expanded from single-clip imitation to learning from large, heterogeneous motion corpora~\cite{gmt}. Behavior foundation models encode that motion diversity in a reusable prior. The bases we adapt in this work span the two dominant forms this takes. SONIC exposes a latent command interface to upstream modules and is trained to track references supplied at run time, so a planner, a teleoperator or a VLA can drive the same controller~\cite{sonic}. Humanoid-GPT instead distills a generalist tracker for zero-shot use on unseen motion, without a task-specific latent~\cite{humanoidgpt}. HoloMotion~\cite{holomotion} and further work on prompting and scaling these priors~\cite{bfmzero,scalingbfm} follow the same interface. Other approaches reach hardware-ready tracking through minimal reward formulations~\cite{beyondmimic}, real-time retargeting for teleoperation~\cite{twist}, and alternative representations of motion priors~\cite{motionpriors}. Accepting a reference from any upstream source is what makes these models reusable as an interface rather than as a skill. Their training distribution, however, is free-space motion on flat ground in empty scenes~\cite{gigabrain}: the policy observes proprioception and a reference window, and the environment enters neither the observation nor the reward. However far the corpus is scaled, foot placement and contact behavior on terrain do not emerge from it.

\subsection{Terrain-aware whole-body control}
Terrain geometry was addressed earlier along the perceptive-locomotion line: elevation-map and depth-based policies traverse stairs, gaps and parkour terrain~\cite{dpl,perceptiveloco,php,extremeparkour}, with separate lines on surefooted placement~\cite{ssr}, on robustness when exteroception degrades~\cite{rpl}, and on sparse footholds~\cite{beamdojo}; related parkour controllers exist on quadrupeds~\cite{puma}. Their command channel, however, is a velocity direction or a navigation goal, which specifies where the base should go rather than what the whole body should do~\cite{gigabrain}. Three routes bring terrain into a reference-driven BFM. GigaBrain-WBC recovers the geometry each motion must have been performed on and trains a world-model tracker on the resulting terrain-paired corpus. Terrain-interactive motion becomes trackable, but the deployed policy reads only proprioception, the previous action and a latent command~\cite{gigabrain}. GRAIL specializes a pretrained BFM for two task families, and its terrain-aware branch fine-tunes the controller together with a height-map encoder~\cite{grail}. Perceptive BFM is the closest to our setting: it keeps the raw kinematic reference as the deployment command and injects terrain features from a torso-depth height grid through an identity-initialized residual path. The tracking prior is fine-tuned with PPO at the same time, the residual acts on the whole-body actor mean, terrain supervision needs references rewritten to conform to the terrain, and the identity gate is a training initialization rather than a deployment switch~\cite{pbfm}. These three routes each pay a price in how perception and tracking are coupled. The deployed policy of GigaBrain-WBC consumes no exteroception, so its tracking does not depend on perception, at the cost of never measuring the geometry in front of it. GRAIL and Perceptive BFM bring perception in, but the terrain skill and the base weights sit in the same parameters, so a degradation in perception costs more than terrain traversal. These routes motivate a complementary question: how much terrain capability can be obtained without modifying the foundation tracker or adding a task-specific perception-and-control architecture? GRAFT studies this adaptation-side trade-off. Its central claim is that a frozen whole-body prior can reduce the task-specific structure required for terrain adaptation, while preserving the upstream reference interface.

\subsection{Residual and parameter-efficient adaptation of a frozen policy}
Learning a residual correction on top of a fixed policy has developed as a separate line. ResMimic conditions whole-body corrections on object and contact state for loco-manipulation~\cite{resmimic}, MOSAIC uses a residual adapter to close an interface gap, explicitly motivated by avoiding the forgetting that continual fine-tuning causes~\cite{mosaic}, and a cerebellum-inspired controller supplies inference-time corrections for recovery from actuator and dynamics faults~\cite{cerebellar}. In all of them the frozen policy remains runnable once the residual is taken away, and ResMimic evaluates that base-only mode as a baseline. GRAFT differs in what conditions the correction and in what it writes: local terrain geometry rather than object, interface or fault state, and the lower body rather than the whole body. We additionally evaluate disabling and re-enabling the adapter within a single continuous simulation run, which none of them reports.

Related approaches project commands onto a learned competence manifold~\cite{cmp}, modify the weights themselves~\cite{lorapolicy,safelora}, adapt a frozen backbone through learned masks~\cite{maskadapt}, or absorb dynamics shifts on a fixed controller~\cite{residualctrl}. Table~\ref{tab:capability} places GRAFT against these lines: it keeps the frozen-base, additive-residual structure, conditions it on terrain geometry, and treats plug-and-play as a deployment property.

\section{Method}
\label{sec:method}

\begin{figure*}[!t]
\centering
\includegraphics[width=\textwidth]{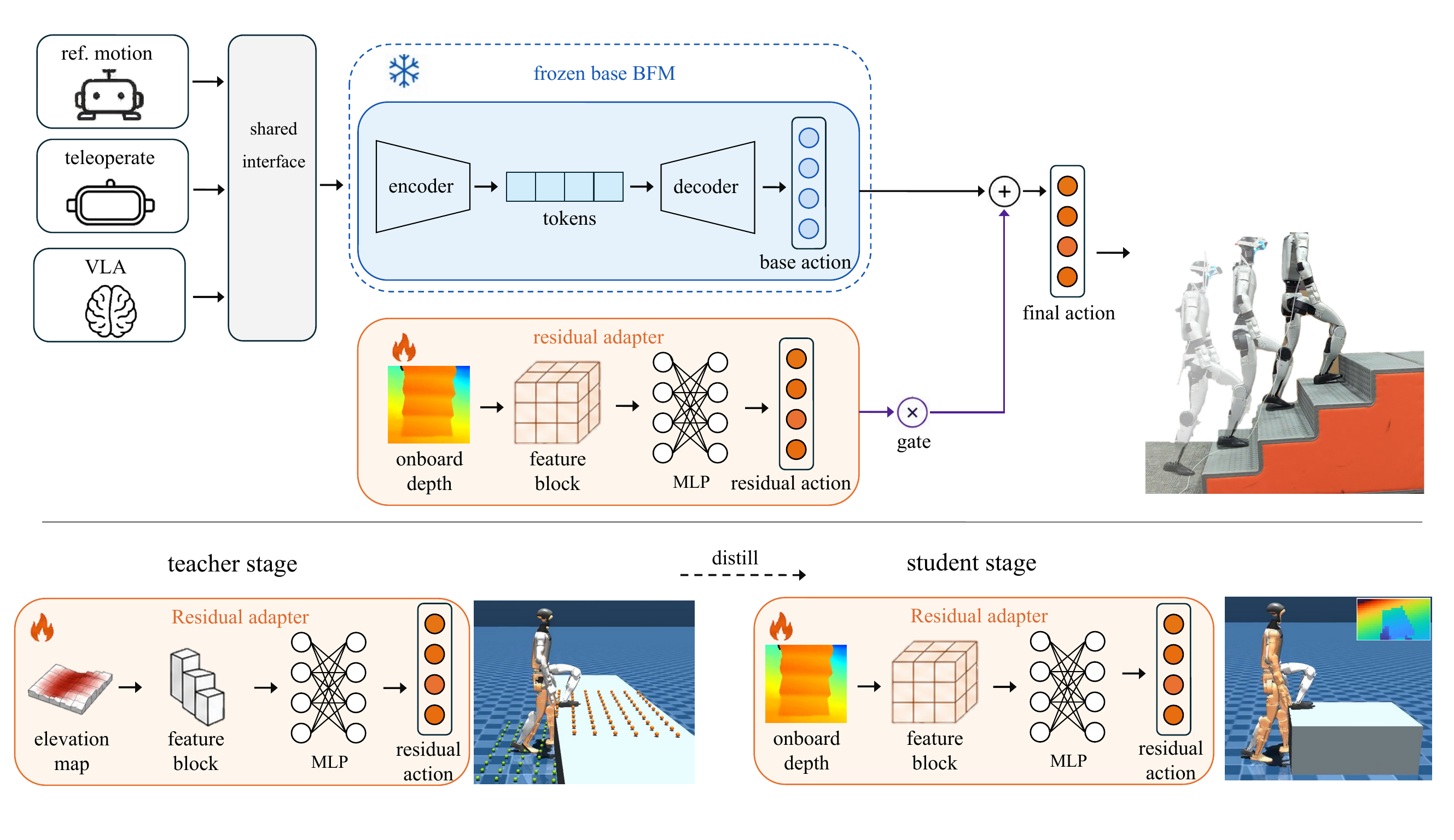}
\caption{GRAFT. Top: the frozen BFM maps the unmodified reference, from a recorded clip, a
teleoperator or a VLA, to a nominal whole-body action, and the adapter reads onboard depth and
adds a correction on the lower-body channels only. Bottom: the adapter is first trained against a
privileged elevation map and then distilled into a student that reads onboard depth. The node marked \emph{gate} closes the
residual pathway at deployment, after which the executed target is the frozen BFM's own output. The adapter's proprioception, base-action and residual-history inputs are omitted for clarity.}
\label{fig:framework}
\end{figure*}

GRAFT separates whole-body motion generation from terrain execution adaptation. The frozen BFM provides a reusable motion prior, including reference tracking, nominal locomotion, and whole-body coordination. A lightweight residual policy then learns the terrain-dependent deviation from this nominal execution. At each control step the frozen BFM $\pi$ maps the kinematic reference and proprioception to nominal whole-body joint-position targets $a^0_t$. A residual policy $f_\phi$ runs at the same control rate and produces corrections in the same units, and the combined targets are executed by a joint-level PD controller,
\begin{equation}
\Delta a_t = m \odot f_\phi(o^{\mathrm{res}}_t),
\qquad
a_t = a^0_t + \Delta a_t,
\end{equation}
where the mask $m$ confines the correction to the lower-body joints. The output of $f_\phi$ carries a fixed per-joint scale and is expressed in the same units as $a^0_t$, so $\Delta a_t$ is the correction actually added. Channels outside the mask are identical to the base output, and zeroing the residual makes the executed target equal $\pi$'s output at every step. Only $\phi$ is trained; $\pi$ stays frozen. The trainable problem is therefore focused on terrain-dependent execution rather than relearning the nominal gait or whole-body reference-following behavior.

The adapter observes proprioception, the nominal base action, its own previous correction and a local terrain observation. It does not receive the reference: what the upstream source asks for reaches the adapter only through that base action, leaving the upstream reference interface unchanged.

At deployment a gate (Fig.~\ref{fig:framework}) selects whether the correction is applied; disabling it leaves $\pi$'s output unchanged, and the gate is set rather than learned. The lower-body mask prevents direct writes to the upper-body targets. It is not an assumption that legs and arms are dynamically independent --- the effect of a leg correction on upper-body tracking is measured in Sec.~\ref{sec:exp}-D.

\subsection{Terrain rewards}
The corrections terrain requires are about where and when a foot arrives, so the reward acts on the feet. Two terms carry most of the terrain behavior and a third suppresses the residual where the ground is flat. They are optimized together with the task reward $r_{\mathrm{task}}$ of the underlying course, which rewards commanded forward progress and staying upright, and with the regularizers $r_{\mathrm{reg}}$ of the base task,
\begin{equation}
r_t = r_{\mathrm{task},t} + r_{\mathrm{reg},t} + \lambda_{\mathrm{clr}}\,r_{\mathrm{clr},t} + \lambda_{\mathrm{fp}}\,r_{\mathrm{fp},t} + \lambda_{\mathrm{flat}}\,r_{\mathrm{flat},t},
\end{equation}
evaluated along rollouts of the combined controller. The three terrain terms are non-positive penalties with non-negative weights. The adaptation-specific recipe is thus limited to foot clearance, foot placement, and flat-ground residual suppression, while the retained base regularizers preserve the released controller's motion and stability conventions. Forward progress and termination are scored whether or not the adapter acts, which is what gives it an incentive to intervene once the terrain makes the base action fail; the terrain penalties then shape the clearance and placement that traversal requires.

\textbf{Swing clearance.} Clearance is scored over the swing rather than at a single instant. At each swing step the terrain observation gives the surface height under the foot, and any deficit against that surface plus a margin $\delta$ is accumulated over the swing and charged at the corresponding touchdown,
\begin{equation}
r^{(i,k)}_{\mathrm{clr}} = -\!\!\sum_{\tau\in\mathcal{S}_{i,k}}\!\! \big[\,h^{\mathrm{surf}}_{i,\tau}+\delta-h^{\mathrm{foot}}_{i,\tau}\,\big]_{+}.
\end{equation}
Here $\mathcal{S}_{i,k}$ is the set of control steps in swing $k$ of foot $i$, $[\,\cdot\,]_{+}$ is the positive part, and foot and terrain heights share one vertical reference.
Scoring the whole swing, rather than the instant of contact, is what encourages the foot to lift ahead of a step instead of into it.

\textbf{Edges and overhang.} The sole of foot $i$ is sampled at a set of footprint points $\mathcal{P}_i$. Only a foot that carries load is charged: $c_{i,t}$ indicates contact, and each of that foot's sample points is penalized when it lies on a visible height discontinuity, indicated by $\mathbf{1}_{\mathrm{edge}}$, and when it overhangs the surface below it beyond a threshold, indicated by $\mathbf{1}_{\mathrm{over}}$,
\begin{equation}
r_{\mathrm{fp},t} = -\sum_{i}\frac{c_{i,t}}{|\mathcal{P}_i|}\sum_{p\in\mathcal{P}_i}
\big(\mathbf{1}_{\mathrm{edge}}(p,t) + \mathbf{1}_{\mathrm{over}}(p,t)\big).
\end{equation}

\textbf{Flat-ground suppression.} A flatness weight is formed from the same terrain observation over two regions, one under the feet and one ahead of the robot,
\begin{equation}
g_t=\exp\!\Big(\!-\tfrac{\mathrm{Var}(h^{\mathrm{u}}_t)+\mathrm{Var}(h^{\mathrm{f}}_t)+(\bar h^{\mathrm{u}}_t-\bar h^{\mathrm{f}}_t)^{2}}{\sigma_h^{2}}\Big),
\end{equation}
where $h^{\mathrm{u}}_t$ and $h^{\mathrm{f}}_t$ are the elevation samples under the feet and ahead of the robot and $\sigma_h$ is a shared length scale. The two variance terms measure flatness within each region; the mean-height difference separates locally flat regions that sit at different elevations, such as the ground and a stair tread. The residual is then suppressed in proportion to this weight, through a bounded surrogate of the weighted residual magnitude,
\begin{equation}
\rho_t = \tfrac{1}{|\mathcal{J}|}\!\sum_{j\in\mathcal{J}}|\Delta a_{t,j}|,
\qquad
r_{\mathrm{flat},t} = -\,g_t\big(1-e^{-\rho_t/\sigma_r}\big),
\end{equation}
with $\rho_t$ the mean absolute correction over the writable lower-body channels $\mathcal{J}$ and $\sigma_r$ its scale. The penalty is bounded between $-g_t$ and zero, which keeps a large residual early in training from dominating the terrain terms. This term shapes behavior during training only; the deployed adapter does not depend on it.

\subsection{Teacher--student training}
\textbf{Stage I: privileged teacher.} The adapter is trained with PPO while the base is frozen, reading a privileged simulated elevation sample as its terrain observation. Episodes terminate when the robot completes the course or meets a failure condition.

\textbf{Stage II: onboard-depth student.} We freeze the teacher and train a student to predict its correction from a rendered onboard-like depth image. The student acts in the environment and is supervised by the teacher under a squared-error loss. At deployment the rendered depth is replaced by the onboard stream, while the residual-action output stays the same. We evaluate whether the student retains the teacher's near-zero flat-ground correction without enforcing it during distillation.

\subsection{Training terrain and references}
Training uses procedurally generated terrain with a difficulty curriculum. The reference motions are drawn from sources whose upper-body motion varies in style and origin, which is what leads the adapter to compensate for the balance effects of a commanded arm motion instead of rewriting it. Domain randomization covers the usual dynamics and observation noise. Instantiating the same procedure on another BFM requires matching the interface it writes against: joint ordering, action units and default posture. What is shared across bases is the residual placement and the objective family; the interface mapping and the base-specific regularizers are adapted, and one adapter is trained per base, so what is reused is the procedure and not the weights.

\section{Experiments}
\label{sec:exp}

We evaluate whether a frozen behavior foundation model can acquire terrain competence and what this adaptation costs:

\begin{itemize}
\item[\textbf{Q1}] Does a frozen whole-body motion prior simplify multi-terrain adaptation, and where should the correction be injected --- on the executed action, on the base's latent command, or in its weights?
\item[\textbf{Q2}] Does the same adaptation procedure apply to different frozen bases?
\item[\textbf{Q3}] How does the adaptation affect flat-ground actions and upper-body tracking, and can it be disabled during a run?
\item[\textbf{Q4}] Does the adapted system work on hardware?
\end{itemize}

\subsection{Setup}
All simulated experiments use a Unitree G1 model. Adapters are trained in Isaac Lab with 4096 parallel environments for 20k iterations at 50~Hz on procedurally generated courses with a difficulty curriculum. We evaluate in MuJoCo on fixed terrains: up stairs, down stairs, a gap, a platform, and flat ground. The reported obstacle results use the hard terrain setting, corresponding to the maximum difficulty used during training. Reference motions are drawn from the SONIC training dataset and are not modified for terrain. Success is the fraction of ten evaluation sessions per cell that reach the terrain end without falling and remain there. We also report the mean $L_2$ norm of the applied lower-body correction in raw action units.

\begin{figure}[t]
\centering
\includegraphics[width=\columnwidth]{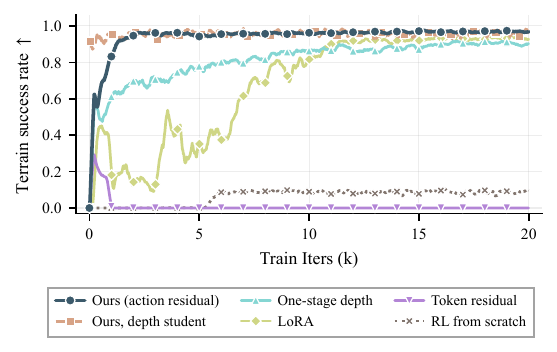}
\caption{Training success on the same frozen SONIC base. Success is averaged over the four obstacle terrains; Table~\ref{tab:routes} reports the corresponding hard-terrain evaluation.}
\label{fig:routes}
\end{figure}

\begin{table}[t]
\centering
\caption{Comparison of terrain adaptation approaches on hard terrain. Each success cell is computed over ten sessions ($n=10$ per cell). Flat residual is reported in raw action units.}
\label{tab:routes}
\scriptsize
\setlength{\tabcolsep}{2pt}
\begin{tabular}{lcccc|c}
\toprule
& \multicolumn{4}{c|}{Success} & Flat resid. \\
Adaptation & Up stairs & Down stairs & Gap & Platform & \\
\midrule
Action residual (ours) & \textbf{1.00} & \textbf{1.00} & \textbf{1.00} & \textbf{0.90} & \textbf{0.0520 $\pm$ 0.0019} \\
Depth student (ours)   & \textbf{1.00} & 0.80 & \textbf{1.00} & \textbf{1.00} & 0.4491 $\pm$ 0.3071 \\
LoRA (decoder)         & \textbf{1.00} & 0.90 & \textbf{1.00} & 0.60 & 0.6929 $\pm$ 0.7173 \\
One-stage depth        & 0.40 & 0.90 & 0.70 & 0.20 & 1.5930 $\pm$ 0.0212 \\
Motion-token residual  & 0.00 & 0.00 & 0.00 & 0.00 & 4.7025 $\pm$ 1.2187 \\
RL from scratch        & 0.00 & 0.10 & 0.00 & 0.00 & -- \\
\bottomrule
\end{tabular}
\end{table}

\subsection{Comparing terrain adaptation approaches}
\label{sec:exp-inject}
We compare six terrain adaptation approaches with the same simulator, reward specification, and iteration budget. The action residual adds a lower-body correction to the executed target; the depth student is its two-stage, onboard-depth version. The motion-token residual perturbs the latent command, LoRA adapts decoder weights, one-stage depth skips the privileged teacher, and RL from scratch uses no base. LoRA represents the fine-tuned terrain-aware BFM route~\cite{grail,pbfm}.

\textbf{(Q1) A residual on the executed action adapts fastest.} As shown in Fig.~\ref{fig:routes}, the action residual reaches 0.947 within 2k iterations and 0.969 by 20k, showing a clear optimization advantage under the same training setup. LoRA reaches 0.925 by 20k and achieves comparable hard-terrain success, but changes flat-ground actions more: $0.6929\pm0.7173$ versus $0.0520\pm0.0019$ for the action residual (Table~\ref{tab:routes}). Updating decoder weights therefore reaches similar terrain performance at the cost of larger and more variable nominal-motion changes.

\textbf{Perturbing the latent command did not produce foothold behavior.} The motion-token residual fails on all four hard obstacle terrains and also disrupts flat-ground behavior, producing the largest flat residual at $4.7025\pm1.2187$. Because the token is a compressed representation of the reference motion and its decoder is trained for flat-ground tracking, perturbing the token changes the decoded motion without providing a mechanism for terrain interaction or foothold adjustment.

\textbf{A frozen motion prior simplifies terrain adaptation.} The BFM already supplies locomotion, reference tracking, and whole-body coordination, so the adapter only has to learn terrain-dependent execution. Under the same evaluation, the action residual achieves 39/40 hard-terrain successes, whereas RL from scratch achieves only 1/40. This gap supports the view that the frozen prior changes the optimization problem from learning locomotion and terrain interaction jointly to learning a focused execution correction.

\textbf{The depth student outperforms direct depth training.} The depth student reaches 0.963 in training and achieves 38/40 successes on hard terrain, although its failures are distributed differently across terrain types. One-stage depth reaches 0.901 by 20k iterations but obtains only 22/40 hard-terrain successes and produces a larger flat residual of $1.5930\pm0.0212$, compared with $0.4491\pm0.3071$ for the distilled student. The result suggests that privileged-to-depth distillation improves both optimization and execution stability.

\begin{figure}[t]
\centering
\includegraphics[width=\columnwidth]{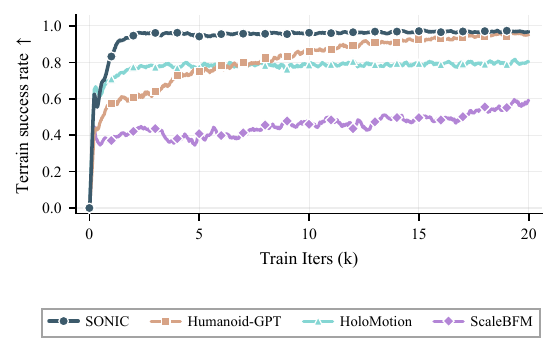}
\caption{Training success for the same adaptation procedure on four frozen bases. Each base has its own adapter; the curves are not a ranking of the trackers. Table~\ref{tab:bfm} reports the corresponding hard-terrain evaluation.}
\label{fig:bfm}
\end{figure}

\begin{table}[t]
\centering
\caption{Cross-base terrain adaptation on hard terrain. Each success cell reports base/GRAFT performance over ten sessions ($n=10$ per cell); flat residual is reported for GRAFT in raw action units. H-GPT and HoloMot. denote Humanoid-GPT and HoloMotion, respectively. All trackers are driven by reference motions from the SONIC training dataset.}
\label{tab:bfm}
\footnotesize
\setlength{\tabcolsep}{2.1pt}
\begin{tabular}{lcccc|c}
\toprule
& \multicolumn{4}{c|}{Success (base/GRAFT)} & Flat resid. \\
Base & Up stairs & Down stairs & Gap & Platform & GRAFT \\
\midrule
SONIC    & 0.00/\textbf{1.00} & 0.30/\textbf{1.00} & 0.00/\textbf{1.00} & 0.00/\textbf{0.90} & 0.0520 $\pm$ 0.0019 \\
H-GPT    & 0.00/\textbf{1.00} & 0.20/\textbf{0.90} & 0.10/\textbf{1.00} & 0.00/0.50 & 0.0781 $\pm$ 0.0021 \\
HoloMot. & 0.00/\textbf{1.00} & 0.80/\textbf{0.90} & 0.00/\textbf{1.00} & 0.00/\textbf{1.00} & 0.0648 $\pm$ 0.0009 \\
ScaleBFM & 0.00/0.30 & 0.10/0.00 & 0.10/\textbf{0.90} & 0.00/\textbf{0.80} & 0.0928 $\pm$ 0.0024 \\
\bottomrule
\end{tabular}
\end{table}

\subsection{Transfer across frozen bases}
\textbf{(Q2) The same adaptation procedure applies to different frozen bases.} Each base receives a separate adapter with the same residual placement and objective family; only interface mappings and base-specific regularizers are adapted. Fig.~\ref{fig:bfm} and Table~\ref{tab:bfm} show the resulting training and hard-terrain success.

The training curves show that the same residual recipe converges on all four bases, but at different rates and final levels. SONIC converges fastest and reaches the highest training success, Humanoid-GPT follows a similar trend, while HoloMotion and especially ScaleBFM converge more slowly or plateau lower. This confirms that the adaptation procedure can be instantiated across bases, while the optimization difficulty remains base-dependent.

The hard-terrain evaluation is consistent with this base dependence. All trackers are driven by reference motions from the SONIC training dataset. Without GRAFT, the four bases obtain 3/40, 3/40, 8/40, and 2/40 successes, respectively. With GRAFT, these totals increase to 39/40, 34/40, 39/40, and 20/40. The gains are substantial overall but are not uniform across terrain cells: ScaleBFM, for example, decreases from 1/10 to 0/10 on down stairs, while its main gains occur on the gap and platform.

\begin{figure}[t]
\centering
\includegraphics[width=\columnwidth]{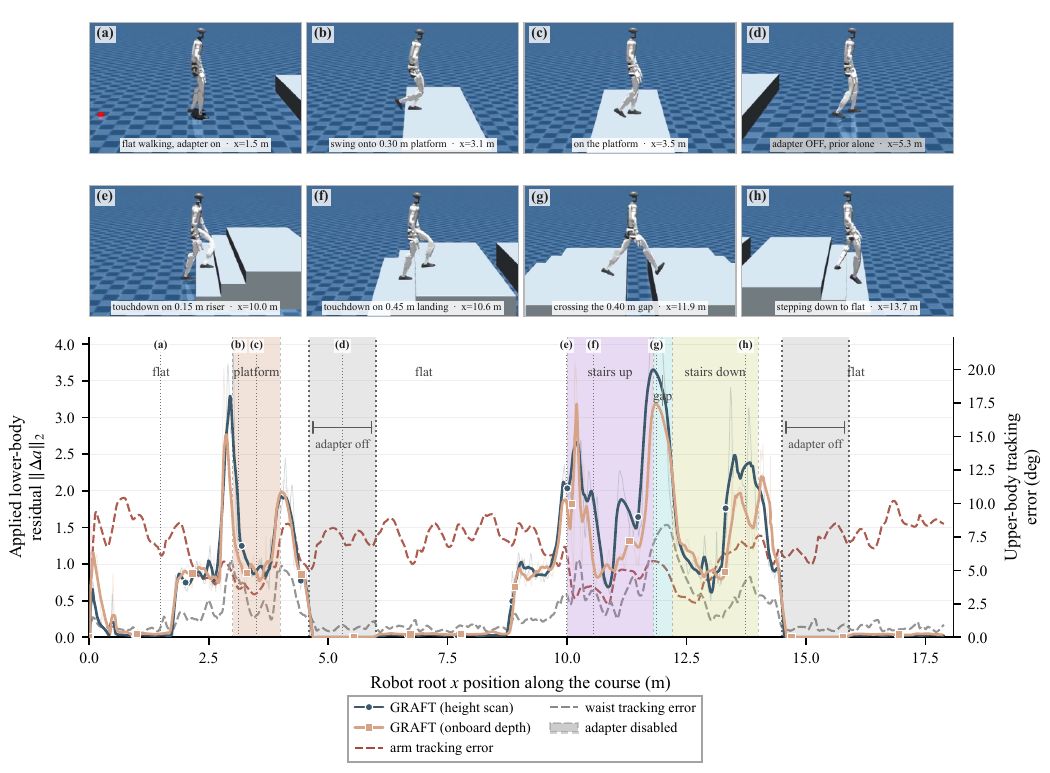}
\caption{Composite-course simulation trace. The residual follows terrain geometry while arm and waist tracking remain visible; grey intervals indicate simulation-only residual disabling.}
\label{fig:trace}
\end{figure}

\subsection{Adaptation cost and removal}
\textbf{(Q3) The correction follows terrain geometry.} Fig.~\ref{fig:trace} shows that, across the composite course, the mean residual is 1.529 on obstacle segments versus 0.408 on flat segments; after removing approach and recovery margins, the values are 1.492 versus 0.184. It peaks at 3.488 on the gap. The depth student shows the same terrain-dependent pattern, while the four-base hard-terrain flat residuals range from $0.0520\pm0.0019$ to $0.0928\pm0.0024$ (Table~\ref{tab:bfm}).

\textbf{Upper-body tracking is largely preserved.} Arm error remains comparable between the adapter-active and adapter-disabled intervals. Waist error rises from 1.18$^\circ$ on flat ground to 3.17$^\circ$ on obstacles, reflecting torso compensation for larger leg motion.

\textbf{In the composite-course simulation, the module can be disabled mid-run.} On flat ground, disabling the gate set the correction to zero while walking continued; re-enabling restored the prior residual level without a transient. This directly demonstrates the switchable base pathway.

\begin{figure}[t]
\centering
\setlength{\tabcolsep}{1pt}
\begin{tabular}{@{}c@{\hspace{2pt}}c@{\hspace{2pt}}c@{}}
\rotatebox{90}{\footnotesize SONIC} &
\includegraphics[width=0.465\columnwidth]{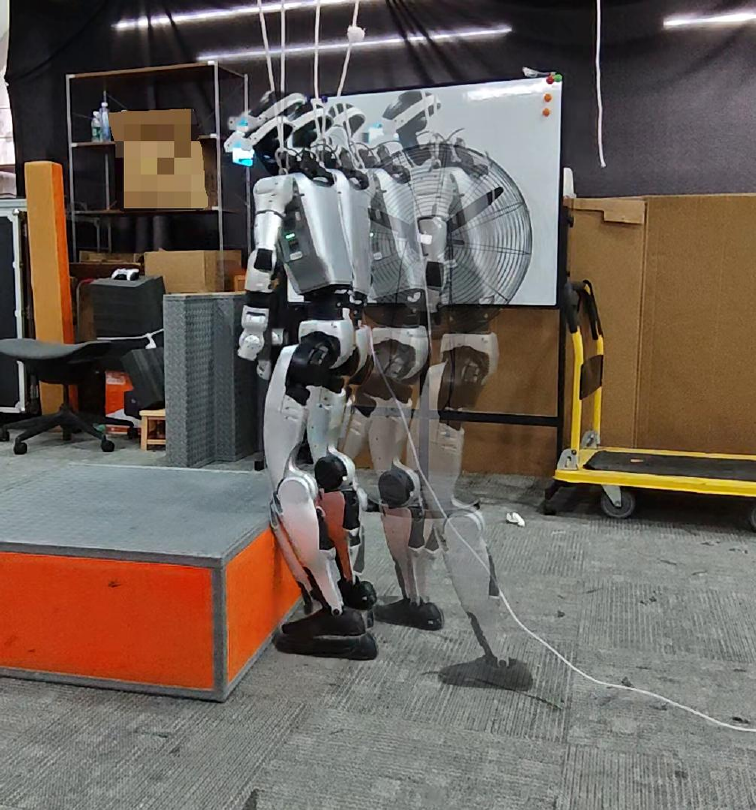} &
\includegraphics[width=0.465\columnwidth]{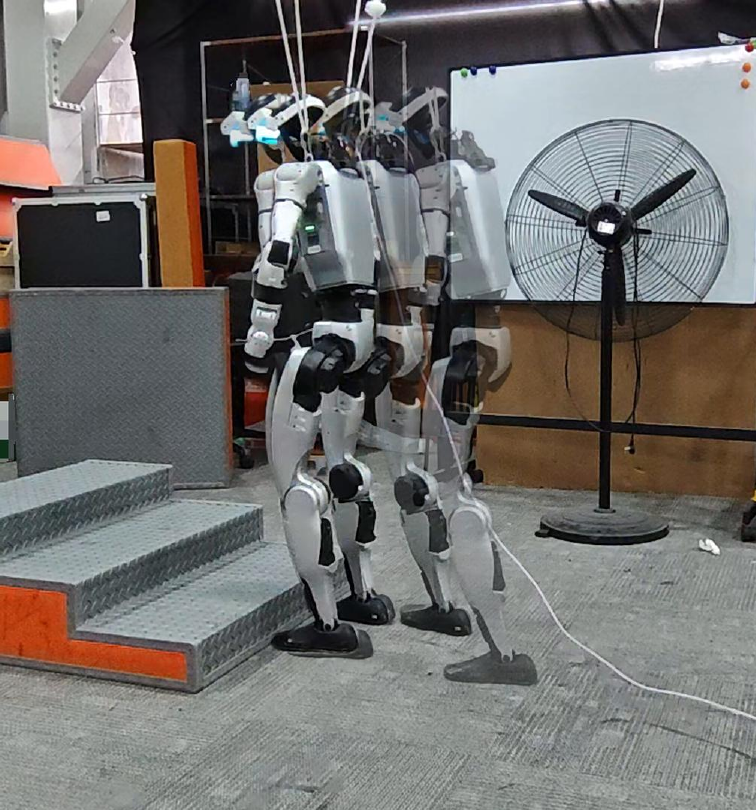} \\[1pt]
\rotatebox{90}{\footnotesize +\,GRAFT} &
\includegraphics[width=0.465\columnwidth]{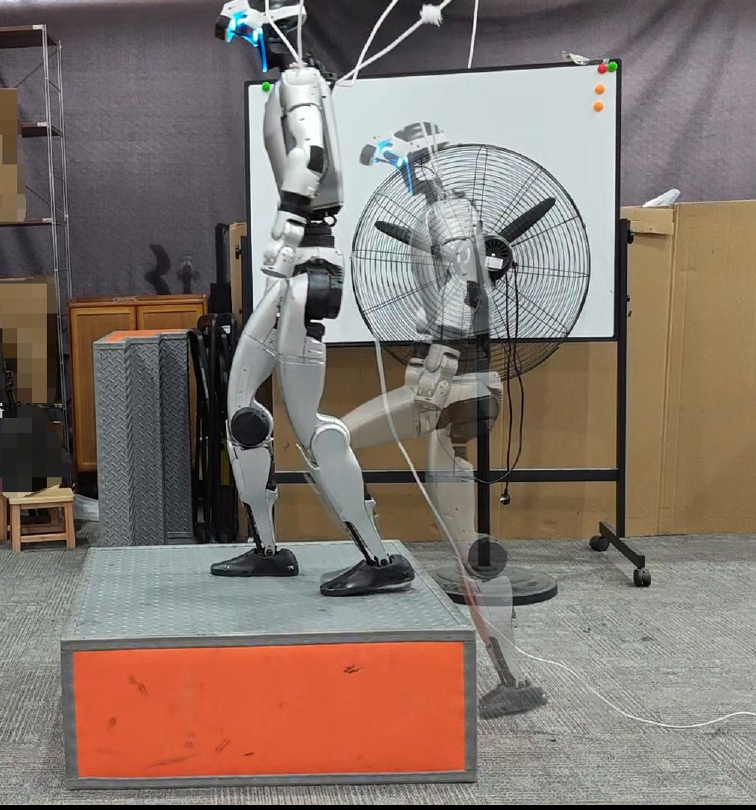} &
\includegraphics[width=0.465\columnwidth]{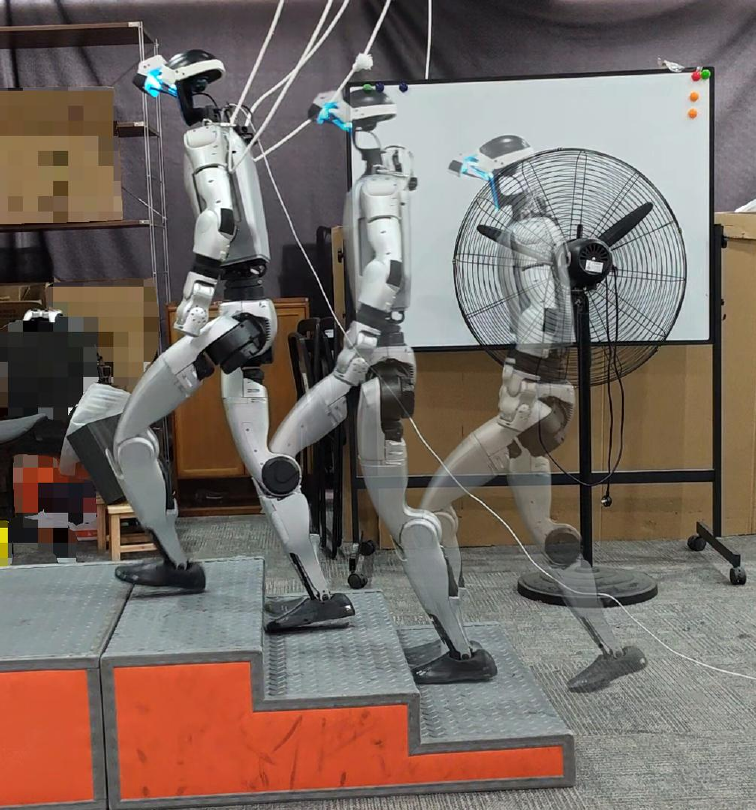} \\[1pt]
& {\footnotesize 0.30\,m platform} & {\footnotesize 0.10\,m step} \\
\end{tabular}
\caption{Hardware comparison with the same base and references. GRAFT raises the foot and crosses both obstacles; the frozen base alone treads in place.}
\label{fig:compare}
\end{figure}

\subsection{Hardware}
\textbf{(Q4) The adapted system works on the robot.} With onboard depth, the adapted SONIC crosses a 0.30~m platform and a 0.10~m step using the same references, whereas the frozen base treads in place before both (Fig.~\ref{fig:compare}).

The same policy also handles a 0.30~m platform, 0.10~m and 0.17~m steps, and a push-cart scene (Fig.~\ref{fig:teaser}), using the released base and unmodified references. The push-cart scene is outside the procedural terrain distribution used during training and serves as an additional qualitative deployment test.

\section{Conclusion}
\label{sec:conclusion}

We presented GRAFT as a separation between a reusable whole-body motion prior and a lightweight terrain-execution adapter. The frozen BFM provides reference tracking, nominal locomotion, and whole-body coordination, allowing the adapter to acquire terrain-dependent foot placement with a compact task-and-foot reward recipe. Across four frozen trackers, this decomposition enabled traversal of stairs, gaps, and platforms that the original policies largely failed to cross, while keeping corrections small on flat ground. In a composite-course simulation, the SONIC adapter could also be disabled and re-enabled during flat-ground walking. Distilling this adapter from privileged terrain observations to onboard depth enabled real-world deployment on a Unitree G1 with unmodified references.

These results show that a frozen whole-body prior can turn multi-terrain learning from a complete locomotion-learning problem into a lightweight execution-adaptation problem. The current evaluation uses recorded motion references and focuses on terrain adaptations achievable through leg corrections. A next step is to evaluate tracking under live teleoperation and VLA commands, particularly when changes in upper-body motion require coordinated adjustments of balance and foot placement.

\addtolength{\textheight}{-11.5cm}

\bibliographystyle{IEEEtran}
\bibliography{refs}

\end{document}